\documentclass[runningheads]{llncs}

\usepackage[T1]{fontenc}
\def\doi#1{\href{https://doi.org/\detokenize{#1}}{\url{https://doi.org/\detokenize{#1}}}}
\usepackage{graphicx}
\usepackage[utf8]{inputenc}
\usepackage{listings}
\usepackage{hyperref}
\usepackage{amsmath}
\usepackage{amsfonts}
\usepackage{booktabs}
\usepackage{placeins}
\usepackage{tabu}
\usepackage{xcolor}

\newcommand{\eg}{\textit{e.g.}}

\usepackage{xcolor}

\begin{document}
\newcommand\ang{66}
\newcommand\angtwo{30}

\title{4D-OR: Semantic Scene Graphs \\for OR Domain Modeling}

\author{Ege Özsoy\inst{1,}\thanks{Both authors share first authorship.}, Evin Pınar Örnek\inst{1,\star}, Ulrich Eck\inst{1}, Tobias Czempiel\inst{1}\\ Federico Tombari\inst{1,2}, Nassir Navab\inst{1,3}}

\authorrunning{E. Özsoy, E. Örnek, U. Eck, T. Czempiel, F. Tombari, N. Navab}

\institute{
Computer Aided Medical Procedures, Technische Universit{\"a}t M{\"u}nchen, Germany
\and
Google
\and
Computer Aided Medical Procedures, Johns Hopkins University, Baltimore, USA 
}

\maketitle 
\begin{abstract}
Surgical procedures are conducted in highly complex operating rooms (OR), comprising different actors, devices, and interactions. To date, only medically trained human experts are capable of understanding all the links and interactions in such a demanding environment. This paper aims to bring the community one step closer to automated, holistic and semantic understanding and modeling of OR domain.  Towards this goal, for the first time, we propose using semantic scene graphs (SSG) to describe and summarize the surgical scene. The nodes of the scene graphs represent different actors and objects in the room, such as medical staff, patients, and medical equipment, whereas edges are the relationships between them. To validate the possibilities of the proposed representation, we create the first publicly available 4D surgical SSG dataset, 4D-OR, containing ten simulated total knee replacement surgeries recorded with six RGB-D sensors in a realistic OR simulation center. 4D-OR includes 6734 frames and is richly annotated with SSGs, human and object poses, and clinical roles. We propose an end-to-end neural network-based SSG generation pipeline, with a rate of success of 0.75 macro F1, indeed being able to infer semantic reasoning in the OR. We further demonstrate the representation power of our scene graphs by using it for the problem of clinical role prediction, where we achieve 0.85 macro F1. The code and dataset will be made available upon acceptance.


\end{abstract}

\section{Introduction}

\begin{figure}[hbt!]
	\centering
	\includegraphics[width=1.0\linewidth]{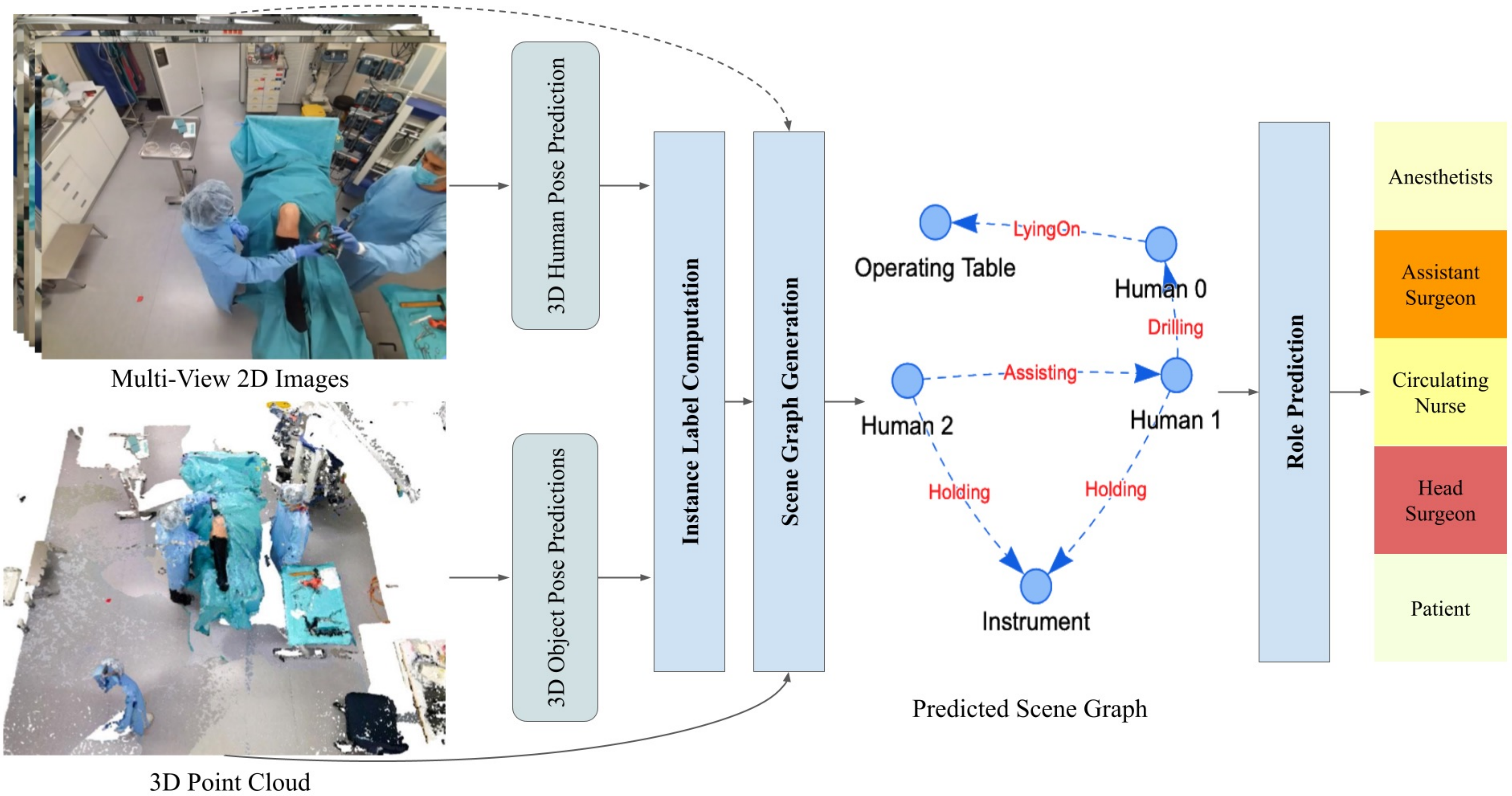}
	\caption{\textbf{An overview of our scene graph generation pipeline.} We predict 3D human poses from images and object bounding boxes from point clouds and assign an instance label to every point. The scene graph generation then uses the fused point cloud, instance labels and images to predict the relations between the nodes, resulting in a semantically rich representation.}
	\label{fig:method_overview}
\end{figure}

An automatic and holistic understanding of all the processes in the operating room (OR) is a mandatory building block for next-generation computer-assisted interventions \cite{maier-hein_sds_2017,kennedymetz2020,mohareri20203d,lalys2014}. ORs are highly variant, unpredictable, and irregular environments with different entities and diverse interactions, making modeling surgical procedures fundamentally challenging.
The main research directions in surgical data science (SDS) focus on the analysis of specific tasks such as surgical phase recognition, tool recognition, or human pose estimation \cite{garrow2020,nwoye2019weakly,czempiel2020,bodenstedt2020,irotoolrecognition2017,mvor}.  
Even though some methods combine different tasks such as surgical phase and tool detection, most previous works operate on a perceptual level, omitting the parallel nature of interactions. To fully understand the processes in an OR, we have to create models capable of disentangling different actors, objects, and their relationships, regarding the OR as one interwoven system rather than multiple individual lines of action. Such models would allow digital systems, \eg, \textit{medical robots, imaging systems}, or \textit{user interfaces}, to automatically adapt their state, function and data visualization, creating an optimized working environment for the surgical staff and improving the final outcome. In the computer vision field, Johnson et al. \cite{image_retrieval_using_scene_graphs} introduced scene graphs as an abstraction layer for images, where nodes represent objects or humans in the scene and edges describe their relationships. Since then, scene graphs have been employed for a variety of tasks such as image editing, caption generation or scene retrieval \cite{image_retrieval_using_scene_graphs,sg2im,dhamo2020,action_genome,3dssg,3d_scene_graph_structure_for_unified_semantics,3d_dynamic_scene_graphs}. The analysis of a scene with scene graphs allows combining different perceptual tasks in one model with holistic knowledge about the entire scene. Most computer vision and scene graph generation methods are designed for everyday-task benchmark datasets \cite{imagenet,scannet,nuscenes,nyu,visual_genome}, where content and interactions are simpler compared to the high complexity in modern OR. Despite the flexible application possibilities, scene graphs have not been used to model the OR specific 3D dynamic environment with complex semantic interactions between different entities.

The learning process of a scene graph generation model for the OR requires a dataset containing annotations for multiple tasks that summarize the events in the scene. Sharghi et al. \cite{surgical_activity_recognition} create a dataset capturing different robot-assisted interventions with two Time-of-Flight video cameras. Their research focuses on the phase recognition of the OR scene on an in-house dataset. Srivastav et al. \cite{mvor} created and published the MVOR dataset, which contains synchronized multi-view frames from real surgeries with human pose annotations. 
This dataset aimed to promote human pose recognition in the operating room and does not provide semantic labels to describe a surgical scene. Currently, no suitable publicly available dataset in the SDS domain exists to supervise the training of a semantic scene graph generation method for the OR.

As we believe in the potential impact of semantic scene graphs for the SDS community, we propose a novel, end-to-end, neural network-based method to generate SSGs for the OR. Our network builds an SSG that is structured, generalizable, and lightweight, summarizing the entire scene with humans, objects, and complex interactions. To this end, we introduce a new 4D dynamic operating room dataset (4D-OR) of simulated knee surgeries, annotated with human and object poses, SSG labels, and clinical roles. 4D-OR is the first publicly available 4D (3D+time) operating room dataset promoting new research directions in SDS. Finally, we demonstrate the capability of our SSG representation by using them for clinical role prediction.

\section{Methodology}

\subsection{Semantic Scene Graphs}
Scene graphs are defined by a set of tuples $G = (N, E)$, with $N = \{n_{1},...,n_{n}\}$ a set of nodes and $E \subseteq N \times R \times N$ a set of edges with relationships $R = \{r_{1},...,r_{M}\} $\cite{image_retrieval_using_scene_graphs}. A 3D scene graph considers the entire 3D representation of an environment. For the OR, the medical equipment and humans are the nodes of our graph, \eg, \textit{anesthesia machine} or \textit{operating table}. The edges describe the interactions between nodes, such as: human \textit{cutting} the patient. 

\subsection{4D-OR Dataset}
To enable and evaluate the modeling of the complex interactions in an OR through SSGs, we propose a novel 4D operating room dataset. 4D-OR consists of ten simulated total knee replacement surgeries captured at a medical simulation center by closely following a typical surgery workflow under the consultancy of medical experts. For the type of intervention, we decided on total knee replacement, a representative orthopedic surgery, which involves different steps and interactions. 4D-OR includes a total of 6734 scenes, recorded by six calibrated RGB-D Kinect sensors \footnote[1]{https://azure.microsoft.com/en-us/services/kinect-dk/} mounted to the ceiling of the OR, with one frame-per-second, providing synchronized RGB and depth images.  4D-OR is the first publicly available and semantically annotated dataset. We provide fused point cloud sequences of entire scenes, automatically annotated human 6D poses and 3D bounding boxes for OR objects. Furthermore, we provide SSG annotations for each step of the surgery together with the clinical roles of all the humans in the scenes, \eg, \textit{nurse, head surgeon, anesthesiologist}. More details are provided in the supplementary material.

\begin{figure}[t!]
	\centering
	\includegraphics[width=1.0\linewidth]{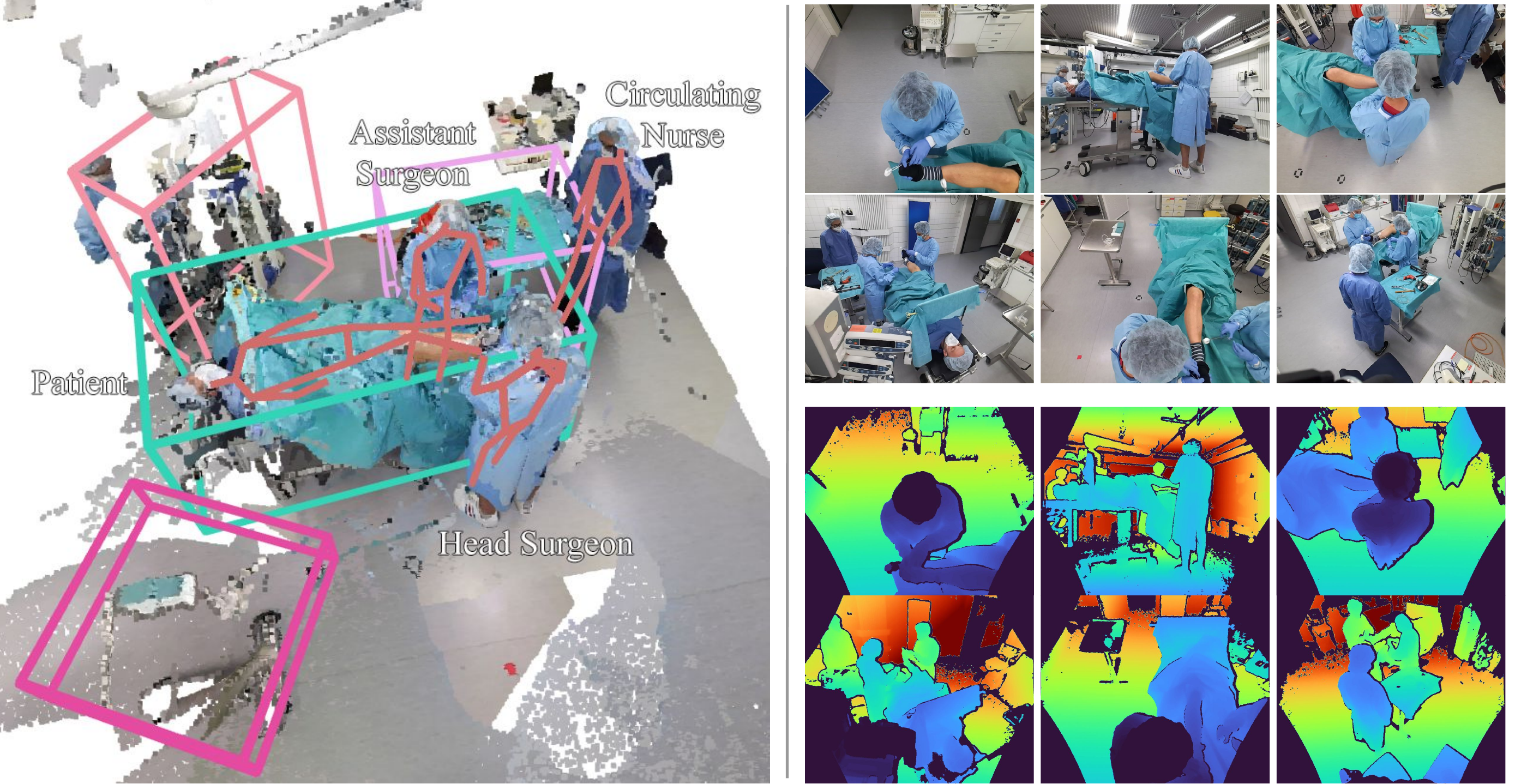}
	\caption{\textbf{A sample 4D-OR scene with multiview RGB-D frames and fused 3D point cloud} with detected 3D object bounding boxes, human poses and clinical roles.}
	\label{fig:example_recognition}
\end{figure}

\subsection{Scene Graph Generation}
Scene graph generation aims at estimating the objects and their semantic relationships from a visual input, \eg, image or point cloud. In this work, we propose an SSG generation pipeline, visualized in Fig. \ref{fig:method_overview}.
Our method detects the humans and objects, extracts their visual descriptors, and builds the scene graph by predicting the pair-wise relations. 
For human and object pose estimation, we use the state-of-the-art methods, VoxelPose \cite{voxelpose} and Group-Free \cite{groupfree}. For scene graph generation, we develop an instance label computation algorithm, which assigns every point in the point cloud to an object instance label using the predicted poses. Additionally, we introduce a virtual node called "instrument", to model interactions of humans with small/invisible medical instruments. Finally, we use a modified version of 3DSSG \cite{3dssg} for generating the scene graphs.

\begin{figure}[t!]
	\centering
	\includegraphics[width=1.0\linewidth]{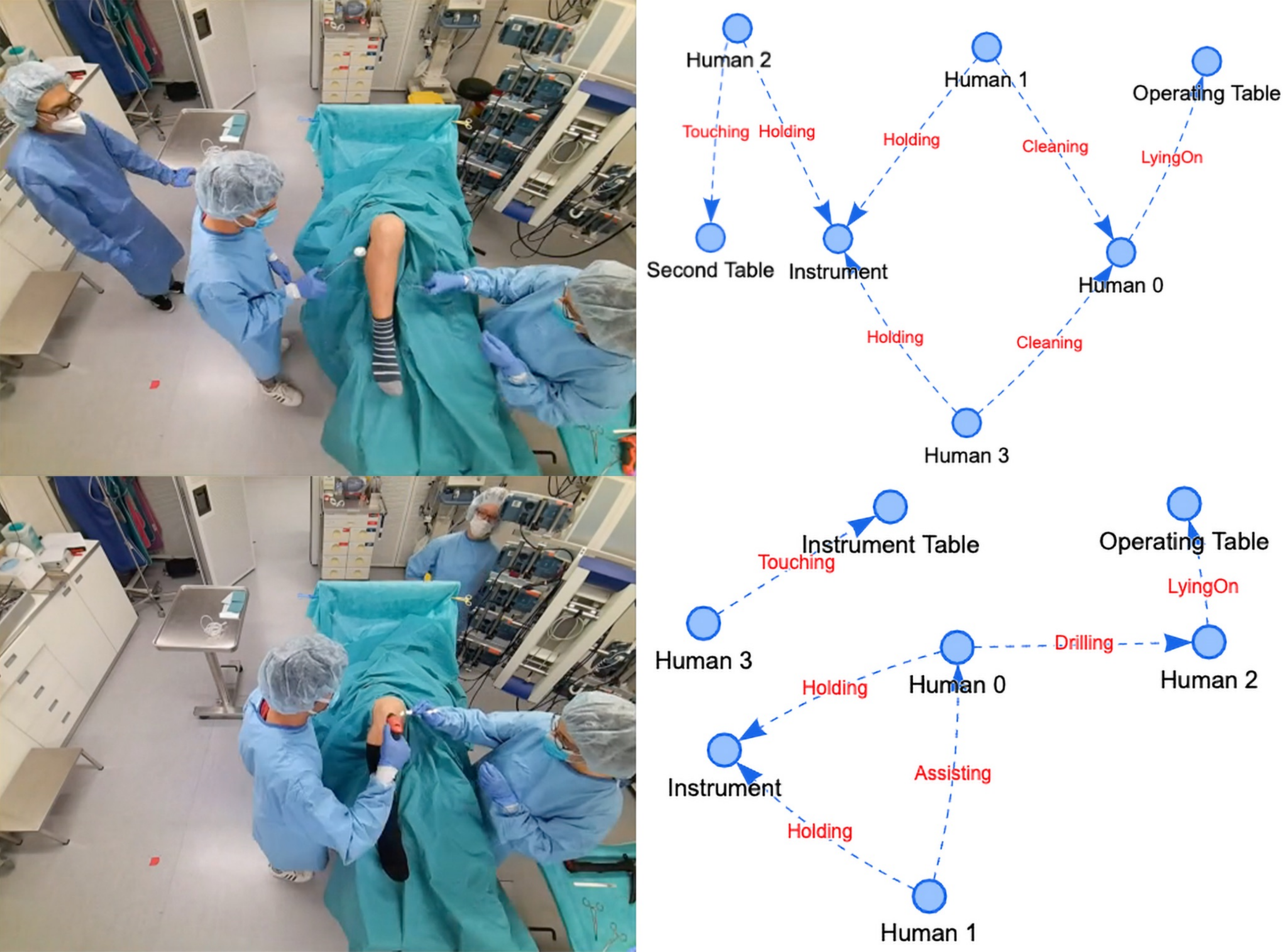}
	\caption{\textbf{Scene graph generation results on two sample scenes.} Visualization from one camera viewpoint, and the scene graph generated from the 3D point cloud.}
	\label{fig:scene_graph_prediction_images_without_closeto}
\end{figure}

3DSSG is a neural network-based approach to predict node relations. As input, it takes a point cloud and the corresponding instance labels for the entire scene. Then, two PointNet \cite{pointnet} based neural networks are used to encode the point clouds into a feature space. The first one, ObjPointNet uses the point clouds extracted at object level. The input to the second one, RelPointNet, is the union of the object pairs' point clouds. Afterward, a graph convolutional network is employed to enhance the features of nodes and edges taking into account the objects and their pair-wise relationships. Finally, the updated representations are processed by multi-layer perceptrons to predict the object and relation classes, trained using the cross-entropy loss. For our scene graph generation method we applied the following domain specific modifications to 3DSSG:

\noindent \textbf{Use of Images:} The OR environment typically contains many objects of various sizes. Especially smaller objects are not always captured well enough by a point cloud \eg, \textit{scissors, lancet}, etc. Further, reflective or transparent items are often omitted due to hardware limitations of depth sensors, even though they can be critical for the correct identification of many OR relations. As these objects are not appropriately represented in point clouds, vanilla 3DSSG often fails to identify them. We propose using images in addition to point clouds to fill this gap. We extract global image features through EfficientNet-B5 \cite{efficientnet} and concatenate them with the PointNet features.

\noindent \textbf{Augmentations:} To simulate real-world variations, such as different shades of clothing, lighting, or sizes, we propose to augment the point clouds by random scale, position, orientation, brightness, and hue during training. For point clouds corresponding to relations, we augment the points of both objects separately to simulate the different sizes or positions of the objects to each other. Finally, we apply a crop-to-hand augmentation, which for relations involving the hands randomly crops the point cloud to the vicinity of the hands. This implicitly teaches the network to focus on medical instruments when differentiating between relations such as \textit{cutting, drilling}, or \textit{sawing}.

\subsection{Use-case: Clinical Role Prediction}
We demonstrate the knowledge encoded in the SSGs in the downstream application of clinical role prediction, which aims to predict the role for every human in the scene. 
To this end, we first associate every human to a track $T$ through a Hungarian matching algorithm using the detected pose at every timestamp. Each track $T$ of length $K$ consists of a subset of generated scene graphs $G_{Ti}$ with $i = \{1,...,K\}$, and a corresponding human node $n_{Ti}$ for that track. To associate the tracks to the human nodes and finally to attain clinical roles, we follow these two steps:

\subsubsection{Track Role Scoring}
For each track $T$, we first compute a role probability score representing the likelihood for each role. For this, we employ a the state-of-the-art self-attention graph neural network Graphormer \cite{graphormer}. We use all the scene graphs within the track $G_T$, and rename nodes $n_{Ti}$ as “TARGET” in corresponding graph $G_{Ti}$, so that the network understands which node embedding to associate the role with. Then, we calculate the mean "TARGET" node embedding and predict the role scores with a linear layer, trained with cross-entropy loss. In addition to this method, we also provide a non-learning baseline for comparison, that is a heuristic-based method which only considers the frequency of relations with respect to each human node. As an example, for every \textit{sawing} relation, we increase the score for \textit{head surgeon}, while for every \textit{lying on} relation, increase the score for \textit{patient}. 

\subsubsection{Unique Role Assignment}
After the track role scoring, we infer the clinical roles of human nodes through solving a matching problem. For every human node in the graph, we retrieve the role probabilities for each track, and bijectively match the roles to the nodes according to their probabilities. Our algorithm guarantees that each human node is assigned to a distinct role.

\section{Experiments}
\noindent \textbf{Model Training}: 
We split the 4D-OR dataset into six train, two validation and two test takes. We configure VoxelPose to use 14 joints, and train it for 20 epochs with a weighted loss for patient poses. We train the 3D object detection network Group-Free for 180 epochs. In our scene graph generation method, we use PointNet++ \cite{pointnet++} for feature extraction, with a balancing loss to deal with rare relations, with 3e-5 lr, 4000 points for object and 8000 for relations. Our method is implemented in PyTorch and trained on a single GPU, achieves a 2.2 FPS inference run-time. 

\noindent \textbf{Evaluation Metrics}: We evaluate human poses with Percentage of Correct Parts (PCP3D), and object poses with average precision (AP) at an intersection over union (IoU) threshold. Further, the scene graph relations and role predictions are evaluated with precision, recall, and F1-score for each class separately and a macro averaged over all classes. Macro averages are unweighted, thus they are sample sizes agnostic, evaluating the performance on all relationships equally.
This is crucial in our setting, as rare relation types such as \textit{cutting} or \textit{drilling} are important for scene understanding. For all metrics, higher means better.

\section{Results and Discussion}
\noindent\textbf{Human and Object Pose Prediction}
In the task of human pose recognition, we reach a PCP3D of 71.23 in the test split. In the task of object pose recognition, we attain an AP value of  0.9893 for IoU@25  and 0.9345 for IoU@50. We further confirmed the reliability of our methods with qualitative results which are visualized in Fig. \ref{fig:example_recognition} through the detected human and object poses.

\noindent\textbf{Scene Graph Generation}
In Tab. \ref{tab:scene_graph_prediction_results}, we present our scene graph generation results in terms of relation prediction from an unlabeled point cloud. We consider a relation \textit{"correct"} if both of the entities are present in the scene and the relation between them is predicted correctly. Overall, we achieve the best result with 0.75 macro F1 using images with point clouds and proposed augmentation strategies. In Fig. \ref{fig:scene_graph_prediction_images_without_closeto}, we visualize two scene graph generation samples qualitatively verifying that our method can indeed understand and reason over semantics in the complex OR and generate correct scene graphs.
For scenes with high visual similarities but different tools (\eg  \textit{drilling, sawing}) our model sometimes fails to predict the correct relation when the instruments are occluded.

\begin{table}[t]
    \caption{\textbf{Quantitative results for scene graph generation} on test split with precision, recall, and F1-score. "Avg" stands for macro average, which is the unweighted average over all classes. We use images, augmentations, linear loss weighting, and PointNet++.}
	\centering
    \begin{tabular}{l|cccccccccccccc|c}
		\rotatebox{\ang}{\textbf{Relation}} & \rotatebox{\ang}{Assist} & \rotatebox{\ang}{Cement} & \rotatebox{\ang}{Clean} & \rotatebox{\ang}{CloseTo} & \rotatebox{\ang}{Cut} & \rotatebox{\ang}{Drill} & \rotatebox{\ang}{Hammer} & \rotatebox{\ang}{Hold} & \rotatebox{\ang}{LyingOn} & \rotatebox{\ang}{Operate} & \rotatebox{\ang}{Prepare} & \rotatebox{\ang}{Saw} & \rotatebox{\ang}{Suture} & \rotatebox{\ang}{Touch} & \rotatebox{\ang}{Avg} \\\midrule
		Prec & 0.42 & 0.78 & 0.53 & 0.97 & 0.49 & 0.87 & 0.71 & 0.55 & 1.00 & 0.55 & 0.62 & 0.69 & 0.60 & 0.41 & 0.68  \\
		Rec & 0.93 & 0.78 & 0.63 & 0.89 & 0.49 & 1.00 & 0.89 & 0.95 & 0.99 & 0.99 & 0.91 & 0.91 & 1.00 & 0.69 & 0.87  \\
		F1 & 0.58 & 0.78 & 0.57 & 0.93 & 0.49 & 0.93 & 0.79 & 0.70 & 0.99 & 0.71 & 0.74 & 0.79 & 0.75 & 0.51 & 0.75  \\
	\end{tabular}
	\label{tab:scene_graph_prediction_results}
\end{table}

\begin{table}[t]
    \centering
  \caption{\textbf{Clinical role prediction results} on test takes, comparing a non-learned heuristic-based method and the neural network based Graphormer for track scoring.}
   \label{tab:role_prediction}
    \begin{tabu} to 0.8\textwidth { X[3l] | X[c] X[c] X[c] | X[c] X[c] X[c]}
         & \multicolumn{3}{|c|}{Heuristic-Based} & \multicolumn{3}{|c}{Graphormer} \\
        \midrule
        Role & Prec & Rec & F1 & Prec & Rec & F1 \\
        \midrule 
        Patient & 0.99 & 0.98 & 0.99 & 0.99 & 0.98 & 0.99 \\
        Head Surgeon & 0.93 & 1.00 & 0.96 & 0.96 & 0.99 & 0.97 \\
        Assistant Surgeon & 0.71 & 0.72 & 0.71 & 0.98 & 0.98 & 0.98 \\
        Circulating Nurse & 0.61 & 0.59 & 0.60 & 0.87 & 0.78 & 0.82 \\
        Anaesthetist & 0.60 & 0.32 & 0.41 & 0.53 & 0.48 & 0.51\\
        \hline
        Macro Avg & 0.77 & 0.72 & 0.74  & \textbf{0.87} & \textbf{0.84} & \textbf{0.85}\\
   \end{tabu}
    \label{tab:baseline_testing}
\end{table}

\noindent\textbf{Clinical Role Prediction}
We present the results on test takes when using heuristics or Graphormer \cite{graphormer} in Tab. \ref{tab:role_prediction}. Example role predictions can be seen in Fig. \ref{fig:example_recognition}. Overall, we achieve near-perfect performance for patient and head surgeon, and good performance for assistant surgeon and circulating nurse. Only the anaesthetist, who is often at least partially occluded, is hard to predict correctly. Unsurprisingly, we observe lower scores with the non-learning heuristic-based score assignment method. While heuristic-based method has the advantage of being transparent, the Graphormer is easier to adapt to new roles or surgeries, as it does not need any tweaking of heuristics.

\noindent\textbf{Ablation Studies} We conduct ablation studies in Tab. \ref{tab:ablation} to show the impact of our two model contributions, using images in addition to point clouds, and augmentations. We also show the impact of relying on ground truth human and object pose annotations instead of predictions. We see that using images (c-d) and augmentations (a-c) significantly improve F1 results, performing the best when both are applied (a-d), confirming the benefits of our method. Further, using ground truth instead of predictions (d-e) does not alter the result much, showing that our method can rely on off-the-shelf pose prediction methods.

\begin{table}[t]
    \centering
    \caption{\textbf{Ablation study} on SSG generation using 3D point clouds with different configurations.}
    \label{ComparisonTable}
    \begin{tabu} to 0.6\textwidth { X[c] X[c] X[c] X[c] | X[c]}
     exp \# & image & augment & GT  & F1  \\ 
        \cmidrule(){1-5}
		 (a) & $\times$ & $\times$ & $\times$  & 0.65 \\
		 (b) & \checkmark & $\times$ & $\times$  & 0.66  \\
		 (c) & $\times$ & \checkmark & $\times$  &  0.70 \\
		 (d) & \checkmark & \checkmark & $\times$  &  0.76 \\
		 (e) & \checkmark & \checkmark & \checkmark  &  0.78 \\
   \end{tabu}
    \label{tab:ablation}
\end{table}

\noindent\textbf{Translation to the clinical setting}
Even if our results are promising, there are multiple obstacles to translating scene graphs into a clinical setting. Data privacy concerns make acquisition, storage, and usage of hospital data challenging. We, however, rely on this rich data source to supervise the proposed method. Nonetheless, in the view of the benefits, we believe that the community will overcome these limitations.

\section{Conclusion}
We propose a semantic scene graph generation method capable of modeling a challenging OR environment. We introduce 4D-OR, a novel scene graph dataset consisting of total knee replacement surgeries to supervise our training. Finally, we demonstrate utilizing generated scene graphs for the exemplary downstream task of clinical role prediction. We believe our work brings the community one step closer to achieving a holistic understanding of the OR.  

\newpage

\bibliographystyle{splncs04}
\bibliography{bibliography}

\end{document}